\documentclass{article}

\usepackage{PRIMEarxiv}

\usepackage[utf8]{inputenc}
\usepackage[T1]{fontenc}
\usepackage{hyperref}
\usepackage{url}
\usepackage{booktabs}
\usepackage{amsfonts}
\usepackage{amsmath}
\usepackage{nicefrac}
\usepackage{microtype}
\usepackage{fancyhdr}
\usepackage{graphicx}
\usepackage{xcolor}
\usepackage{natbib}
\usepackage{algorithm}
\usepackage{algorithmic}
\usepackage{pgfplots}
\pgfplotsset{compat=1.18}
\graphicspath{{media/}}

\definecolor{darkblue}{rgb}{0, 0, 0.5}
\hypersetup{colorlinks=true, citecolor=darkblue, linkcolor=darkblue, urlcolor=darkblue, hypertexnames=false}

\title{Inference-Time Code Selection via Symbolic Equivalence Partitioning}

\author{
David Cho \\
Department of Computer Science \\
Purdue University \\
West Lafayette, IN, USA \\
\texttt{cho353@purdue.edu}
\And
Yifan Wang \\
Department of Computer Science \\
Purdue University \\
West Lafayette, IN, USA \\
\texttt{wang5617@purdue.edu}
\AND
Fanping Sui \\
Texas Instruments \\
Dallas, TX, USA \\
\texttt{f-sui@ti.com}
\AND
Ananth Grama \\
Department of Computer Science \\
Purdue University \\
West Lafayette, IN, USA \\
\texttt{ayg@purdue.edu}
}

\begin{document}
\maketitle

\begin{abstract}
Sampling multiple candidate programs at inference time is an effective way to
improve LLM code generation. However, its benefit depends on reliably selecting
a correct solution from the generated pool. We observe that this selection
problem has a distinctive semantic structure: correct solutions, despite
differences in syntax, implementation, or algorithmic strategy, often converge
to the same functional behavior over valid inputs. At the same time, consensus
alone is not sufficient for correctness, because models can also produce
correlated wrong solutions that implement the same mistaken behavior. We propose
Symbolic Equivalence Partitioning (SEP), an inference-time selection framework
that first uses problem-provided public examples as lightweight validity
signals. SEP then uses symbolic execution to partition the remaining candidate
programs into bounded functional equivalence classes and selects from the dominant
equivalence class. Across HumanEval+ and LiveCodeBench, SEP
consistently improves selection accuracy without auxiliary test generation,
learned verifiers, or additional LLM inference. At $N=10$, SEP improves average
accuracy from 0.754 to 0.826 on HumanEval+ and from 0.565 to 0.647 on
LiveCodeBench, showing that symbolic functional agreement is an
effective signal for inference-time code selection.
\end{abstract}

\section{Introduction} \label{sec:intro}

Large Language Models (LLMs) have demonstrated remarkable capabilities across a wide range of tasks, from text summarization to complex reasoning and code generation \cite{claude26, chen2021evaluating, openai2023gpt4}. To further enhance these capabilities without the prohibitive cost of retraining, inference-time scaling, which allocates additional computational resources during inference, has emerged as a critical paradigm for complex reasoning tasks \cite{openai2024o1, brown2024large}. Recent studies demonstrate that for logic and code, scaling test-time compute often yields higher marginal returns than scaling model parameters \cite{snell2024scaling}. In the domain of code generation, this is commonly realized through ``Best-of-$N$'' methods, where the model generates multiple candidate solutions and a selection mechanism identifies the most promising one.

The effectiveness of Best-of-$N$ methods depends almost entirely on the final selection mechanism. Ideally, correctness would be determined by executing each candidate against a comprehensive suite of ground-truth tests. In practice, such tests are expensive to construct and often unavailable at inference time. Existing selection methods therefore rely on proxy signals. Some methods use LLM-generated test cases to induce execution-based agreement \cite{chen2023codet, chen2025revisit, to-etal-2024-functional}, while others use learned reward models or verifiers to score candidates \cite{ma2024eureka, li2025codeprm}. These approaches can be effective, but they often require auxiliary test generation, additional model inference, verifier training, or larger sample budgets, increasing inference-time cost and latency \cite{li2022competition, brown2024large}.

A common principle behind many Best-of-$N$ selectors is agreement. Prior work shows that correct solutions are often present in a generated candidate pool, even when greedy decoding fails. Methods such as Self-Consistency \cite{wang2022self} and execution-based rerankers such as CodeT \cite{chen2023codet} and SRank \cite{to-etal-2024-functional} exploit the intuition that independently sampled correct solutions are more likely to converge on a shared answer or behavior, whereas incorrect solutions tend to exhibit weaker agreement. However, agreement alone is not a correctness guarantee. When a model has a confident but incorrect interpretation of a problem, independently sampled solutions may collapse onto the same wrong behavior. In such cases, semantic agreement correctly identifies that the wrong programs are equivalent to one another, but majority selection may still choose the wrong behavior.

In this paper, we propose \textbf{Symbolic Equivalence Partitioning (SEP)}, an inference-time selection framework for LLM-generated code. SEP first uses problem-provided public examples, when available, as a lightweight grounding filter. Candidates that fail these examples are removed before symbolic analysis, reducing the chance that a large correlated-wrong group dominates selection. SEP then applies bounded symbolic execution to partition the surviving candidates by functional behavior over valid inputs and selects a representative from the dominant partition. When explicit input-domain constraints are available, SEP injects them as solver-side assumptions to focus symbolic search on valid inputs.

Integrating these techniques into a unified framework, we make the following contributions:
\begin{itemize}
    \item We introduce \textbf{Symbolic Equivalence Partitioning (SEP)}, an inference-time selection framework that filters candidates using problem-provided public examples, applies bounded symbolic execution with optional domain constraints added as solver-side assumptions, and selects from the dominant behavioral partition.
    \item We evaluate SEP on HumanEval+ and LiveCodeBench across seven models and three sampling budgets, showing strong gains over Pass@1 and competitive reranking baselines without requiring auxiliary test generation, learned verifiers, or additional LLM inference beyond the initial $N$ candidate generations.
\end{itemize}

\begin{figure*}[t]
    \centering
    \includegraphics[width=\textwidth]{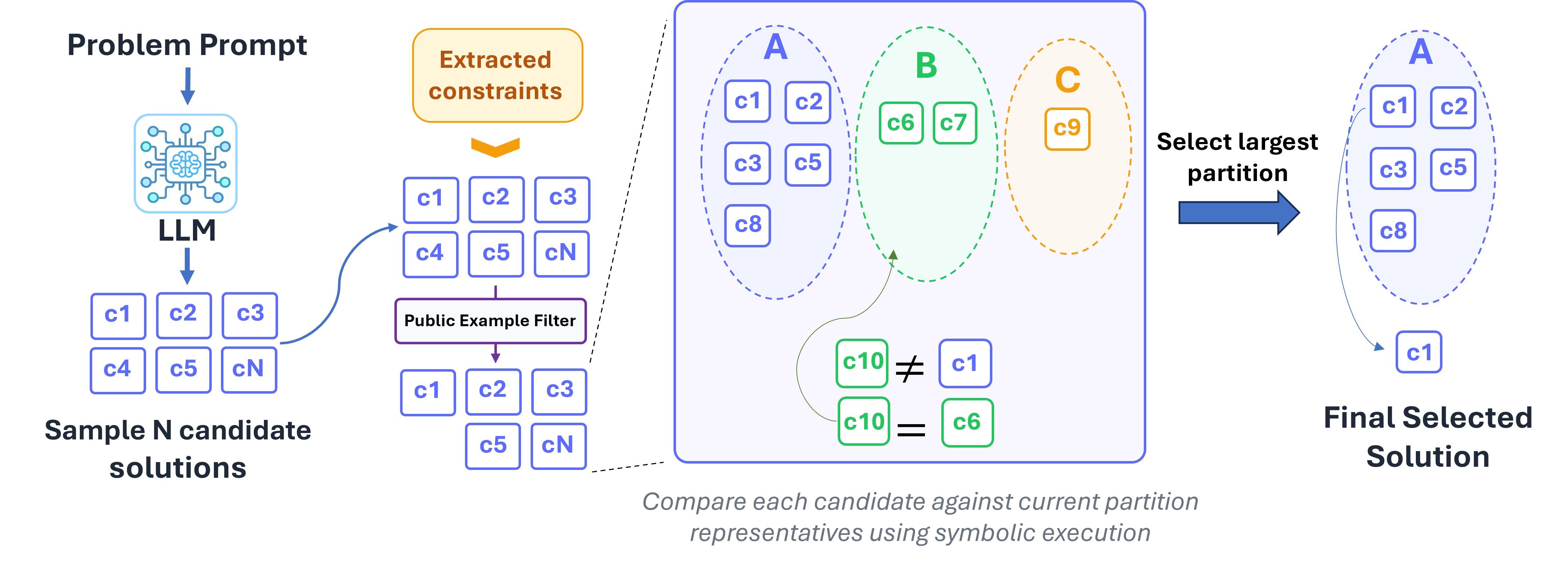}
    \caption{Overview of Symbolic Equivalence Partitioning (SEP).}
    \label{fig:sep_overview}
\end{figure*}

\section{Related Work}
\label{sec:related}

\textbf{Inference-time selection for LLM code generation.}
Best-of-$N$ methods improve code generation by sampling multiple candidate programs and selecting one final answer \cite{chen2021evaluating, cobbe2021training}. Prior selection methods rely on surface agreement, generated tests, or learned verifiers \cite{wang2022self, chen2023codet, to-etal-2024-functional, ma2024eureka, li2025codeprm}. Textual consensus can fail because syntactically different programs may be functionally equivalent, while execution-based methods such as CodeT and SRank use auxiliary generated tests to induce behavioral agreement \cite{chen2023codet, to-etal-2024-functional}. Learned reward models and verifier-based approaches provide another selection signal, but introduce additional training requirements and verifier error \cite{ma2024eureka, li2025codeprm}.

\textbf{Symbolic execution.}
Symbolic execution runs programs on symbolic rather than concrete inputs, maintaining path conditions and querying an SMT solver to prove reachability or produce counterexamples \cite{king1976symbolic, moura2008z3}. It is widely used for testing and bug finding, but often suffers from path explosion as branches and loops multiply the number of explored states. Most related to our work is \citet{sharma2025assessing}, which also uses symbolic analysis for LLM-generated code. The key difference is objective: their method is designed primarily for abstention, whereas SEP uses symbolic equivalence partitioning as a Best-of-$N$ selector. SEP further combines a public example filter with SMT-Constrained Pruning over typed function-level inputs. The filter removes candidates already known to violate the specification, while SMT-Constrained Pruning focuses counterexample search on valid structured inputs under bounded symbolic analysis.

\section{Methodology} \label{sec:methodology}

We propose a \textbf{Symbolic Equivalence Partitioning} framework. Let $s$ denote a problem specification consisting of a natural language description $D$ and, when available, a set of domain constraints $\mathcal{C}$ (e.g., input bounds, value ranges, and structural restrictions). If a problem statement does not provide such constraints, we treat $\mathcal{C}$ as vacuous. An LLM generates a set of $N$ candidate programs $\mathcal{P}_0 = \{p_1, p_2, \dots, p_N\}$ for problem $s$. SEP first applies the public example filter to remove candidates that fail the problem-provided input-output cases. Let $\mathcal{P}$ denote the resulting filtered candidate pool, with cardinality $|\mathcal{P}| = N' \le N$.

We define two programs $p_i$ and $p_j$ as \textbf{functionally equivalent} with respect to domain $\mathcal{C}$, denoted $p_i \equiv_{\mathcal{C}} p_j$, if and only if they exhibit identical execution outcomes for all valid inputs:

\begin{equation} p_i \equiv_{\mathcal{C}} p_j \iff \forall \mathbf{x} \in \mathrm{Dom}(\mathcal{C}), \quad \mathrm{EO}_i(\mathbf{x}) = \mathrm{EO}_j(\mathbf{x}).
 \end{equation}

Our goal is to partition $\mathcal{P}$ into equivalence classes (partitions) $\{K_1, \dots, K_M\}$ and select a representative $p^*$ from the largest partition, under the hypothesis that consensus among diverse correct solutions converges to the ground truth.

\subsection{Public Example Filter}
SEP first executes each candidate on the public example input-output cases provided in the problem statement, when available. Candidates that raise runtime errors or produce outputs inconsistent with the official examples are removed before symbolic partitioning. If no public examples are available, SEP proceeds directly to symbolic partitioning over the original candidate pool.

Although public examples are typically few and far from comprehensive, they serve as an LLM-independent grounding signal that removes candidates already known to violate the specification. As shown in Section~\ref{sec:analysis}, this filtering step substantially reduces the number of incorrect candidates that can form or join the dominant partition.

\subsection{Symbolic Equivalence Checking} \label{sec:diffbehavior}

Verifying $p_i \equiv_{\mathcal{C}} p_j$ can be reduced to checking the unsatisfiability of a divergence formula. We use counterexample-driven symbolic execution to search for an input $\mathbf{x}$ that satisfies both the program's path condition and the domain constraints, such that execution outcomes differ:

\begin{equation}
\exists \mathbf{x} \text{ s.t. } \mathcal{C}(\mathbf{x}) \land \mathrm{PC}(\mathbf{x}) \land \Big(\mathrm{EO}_i(\mathbf{x}) \neq \mathrm{EO}_j(\mathbf{x})\Big)
\label{eq:divergence}
\end{equation}

Here, $\mathcal{C}(\mathbf{x})$ represents domain-specific constraints when they are explicitly available and otherwise defaults to a tautology, while $\mathrm{PC}(\mathbf{x})$ denotes the symbolic path condition (accumulated branch predicates). $\mathrm{EO}_k(\mathbf{x})$ denotes the execution outcome of program $p_k$ on input $\mathbf{x}$, including its return value, any exceptional termination, and any mutations to reachable input objects.

If the SMT solver finds Eq.~\ref{eq:divergence} satisfiable, it yields a counterexample $\mathbf{x}$ witnessing $p_i \not\equiv_{\mathcal{C}} p_j$. If no counterexample is found within the predefined symbolic execution budget, we treat the pair as equivalent for the purposes of partitioning. This yields a bounded approximation to functional equivalence, defined by the explored symbolic paths and resource limits. We discuss this approximation in Section~\ref{sec:limitations}.
In practice, we instantiate this equivalence check using CrossHair \cite{crosshair}, a differential symbolic execution engine for Python that translates program paths into constraints for the Z3 SMT solver \cite{moura2008z3}. All equivalence checks are performed under a fixed resource budget: a per-condition timeout of 15 seconds and a per-path timeout of $\sqrt{15}$ seconds ($\approx 3.87$s). These values are held fixed across benchmarks, models, and candidate pool sizes.

\subsection{SMT-Constrained Pruning}
\label{sec:smt_constraints}

A challenge in applying symbolic execution to competitive-programming problems is that bounded symbolic search may explore inputs outside the problem's valid domain. When explicit input bounds are available but not enforced, the symbolic engine may produce spurious counterexamples or spend symbolic-analysis budget on irrelevant inputs. A simple approach is to insert \texttt{assert()} statements to enforce such constraints; however, in symbolic execution, assertions introduce explicit failure branches that create additional exception paths, often wasting solver budget and hindering effective analysis.
We address this with \textbf{SMT-Constrained Pruning}, a technique that uses \emph{constraint injection} (CI) to encode domain constraints from the problem statement as solver-level SMT assumptions rather than program-level assertions.

\textbf{1. Constraint Extraction and Normalization.} When the problem description contains an explicit constraint block, we extract it from $D$. Constraint expressions involving input parameters are then parsed into Python abstract syntax trees (ASTs) and normalized using syntactic rewrite rules to handle common patterns such as chained comparisons. Benchmarks without explicit blocks, such as HumanEval+, run without constraint injection.

\textbf{2. Injection of Z3 Assumptions.} For every candidate program $p_k$ with extracted constraints, we inject a \textit{symbolic preamble} before the main logic. This preamble adds the extracted constraints directly to the underlying Z3 solver instance. Symbolic paths that violate these assumptions become infeasible and are pruned from the analysis. This constrains the symbolic engine to focus on the valid input domain $\mathrm{Dom}(\mathcal{C})$, reducing invalid input exploration and limiting spurious counterexamples from out-of-domain inputs.

\subsection{Greedy Representative Partitioning} \label{sec:partition}

We implement partitioning using a greedy representative-based procedure. Candidates are processed sequentially and compared against the representatives of existing partitions, ordered from largest to smallest. If symbolic equivalence under $\mathcal{C}$ is established, the candidate joins that partition; otherwise, it initiates a new partition. The final prediction is the representative of the largest partition. Detailed pseudocode is provided in Algorithm~\ref{alg:partition}.
To avoid the cost of performing pairwise comparisons for all $O((N')^2)$ candidate pairs in the post-filter pool, we treat symbolic equivalence as approximately transitive under bounded analysis. This reduces the comparison cost to $O(N' \cdot M)$, where $M$ is the number of distinct semantic partitions, with $M \le N'$.

\subsection{Illustrative SEP Walkthrough}
\label{sec:walkthrough}

To illustrate how SEP constructs semantic partitions in practice, we walk through one LiveCodeBench example, \textit{maximum-strength-of-a-group}, using $N=10$ sampled candidates. In this example, all 10 generated programs pass the public test cases, so the remaining selection problem is to determine which candidates are behaviorally equivalent beyond those examples.

\paragraph{Constraint formation.}
For this problem, SEP extracts the domain constraints:
\[
1 \le |nums| \le 13, \qquad -9 \le nums[i] \le 9.
\]
These constraints are compiled into solver-side assumptions inside the normalized wrapper before symbolic comparison. This keeps symbolic search inside the valid problem domain rather than introducing additional failure branches through program-level assertions.

\paragraph{Bounded symbolic search.}
Each symbolic comparison is performed under a fixed search budget. Internally, the symbolic engine runs the two candidate programs on shared symbolic inputs and queries the SMT solver for a feasible assignment that causes their execution outcomes to differ. This search is deliberately bounded by a per-condition timeout, which limits the total wall-clock time spent exploring paths for one directional comparison, and a per-path timeout, which limits the time allocated to any single symbolic path exploration.

\begin{figure}[t]
    \centering
    \includegraphics[width=0.8\linewidth]{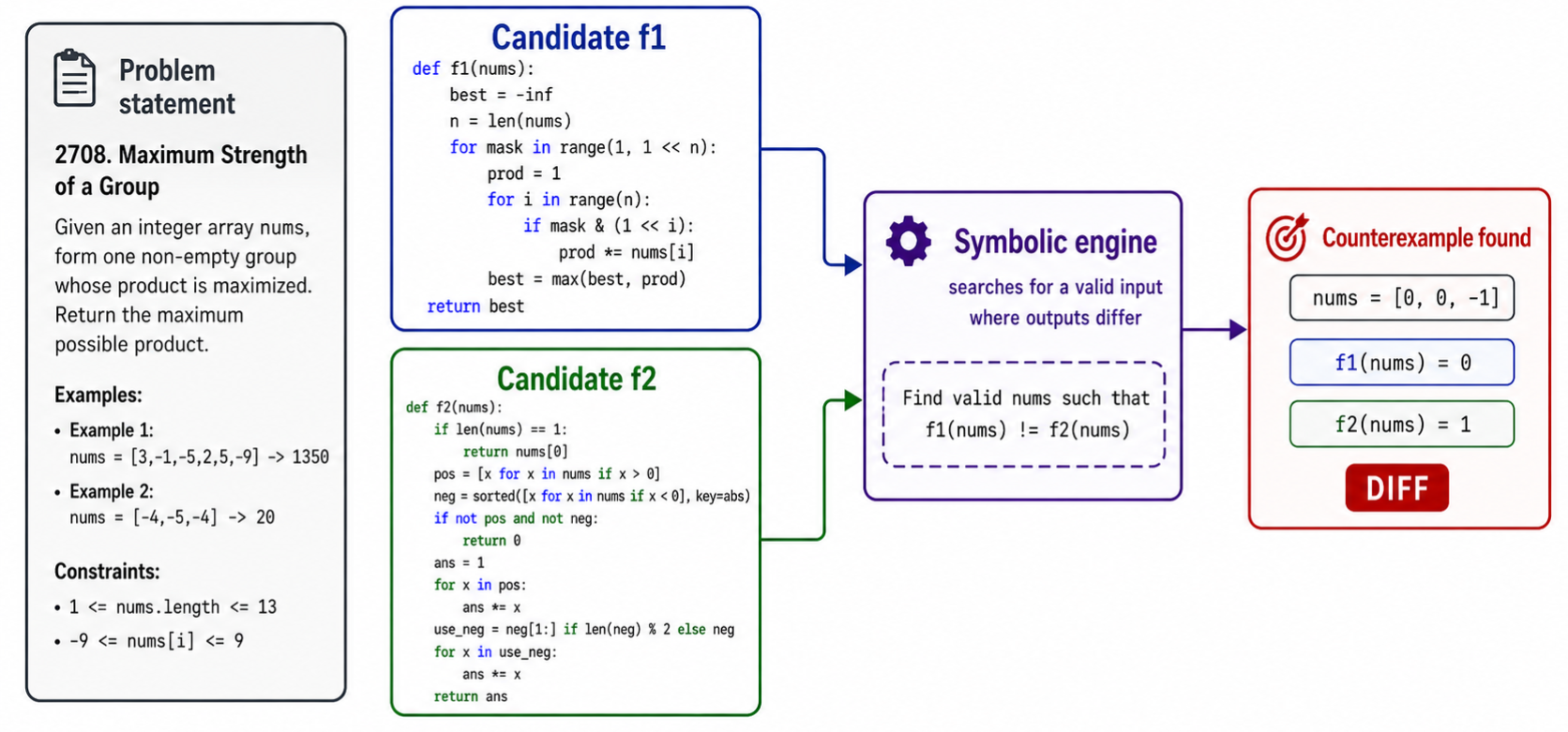}
    \caption{Illustration of SEP separating \texttt{f1} and \texttt{f2} in the walkthrough example. The symbolic engine searches for a valid input satisfying the extracted constraints on which the two programs produce different outputs, and returns a concrete counterexample when such an input is found.}
    \label{fig:symbolic_f1_f2_example}
\end{figure}

\paragraph{Comparisons and partition creation.}
SEP starts with \texttt{f1} as the representative of the first partition. It then compares \texttt{f2} against \texttt{f1}. The symbolic engine finds a valid counterexample,
\[
nums = [0,0,-1],
\]
for which \texttt{f2} returns $1$ while \texttt{f1} returns $0$. SEP therefore marks the pair as \textsc{DIFF} and creates a new partition for \texttt{f2}. Figure~\ref{fig:symbolic_f1_f2_example} illustrates this comparison.

The subsequent candidates go through the same comparison loop against the current partition representatives. Whenever the symbolic engine finds a distinguishing input, SEP places the candidate into a different partition; when no such input is found within budget, SEP merges it into an existing one. In this example, \texttt{f3} and \texttt{f4} are likewise separated into different partitions through the same process, so we omit their remaining concrete witnesses for brevity. By contrast, when \texttt{f5} is compared against representative \texttt{f1}, the symbolic engine reports that no distinguishing input is found within budget. SEP therefore treats \texttt{f5} as equivalent to \texttt{f1} and adds it to the \texttt{f1} partition. The remaining candidates, \texttt{f6, f7, f8, f9, f10}, behave similarly and are also merged into the \texttt{f1} partition.

\paragraph{Resulting partition structure.}
After all 10 candidates are processed in this way, SEP produces four semantic partitions. The dominant partition is represented by \texttt{f1} and contains 7 candidates; all 7 are correct under the benchmark evaluation. The remaining 3 candidates are separated into smaller partitions. This example shows the mechanism SEP is designed to exploit: valid counterexamples create new partitions, while failure to find a differentiating input within the symbolic budget causes candidates to join an existing one. In this instance, the correct solutions concentrate in one large partition, while divergent behaviors remain fragmented. The representative of the largest partition is selected as the final answer.

\section{Experiments} \label{sec:experiments}

\subsection{Experimental Setup} \label{sec:setup}

We evaluate our Symbolic Equivalence Partitioning framework on competitive code generation, comparing it against textual, execution-based, and model-based reranking baselines.

\subsubsection{Benchmarks} \label{sec:benchmarks}

We evaluate on two complementary code-generation benchmarks: HumanEval+ \cite{liu2023evalplus}, a set of 164 Python function-synthesis problems with an augmented hidden test suite, and LiveCodeBench \cite{jain2024livecodebench}, a contamination-resistant benchmark of recent contest-style programming problems. We evaluate on 438 filtered LiveCodeBench problems after excluding tasks that lack a function-level interface or contain incorrect or inconsistent ground-truth tests. These choices are made solely to support function-level symbolic analysis and reliable evaluation. Appendix~\ref{app:benchmark_details} provides additional benchmark implementation details, including our HumanEval+ signature standardization procedure and the excluded LiveCodeBench problems.

\subsubsection{Models} \label{sec:models}
We evaluate our method across a diverse set of large language models spanning 4B to 20B parameters, including both instruction-tuned and reasoning-specialized variants. Specifically, we evaluate the Phi-4 family (Mini-Instruct and Reasoning) and the Qwen3 family (4B, 4B-Instruct-2507, and 14B), alongside Llama-3.1-8B-Instruct and GPT-OSS-20B.

For each model, we use the default or officially recommended sampling parameters when available, and otherwise use settings from a closely related model in the same family. Exact sampling configurations are reported in Appendix~\ref{app:param}.

\subsubsection{Baselines \& Metrics} \label{sec:baselines}

We compare SEP against Pass@1, Pass@$N$, CodeT \cite{chen2023codet}, SRank \cite{to-etal-2024-functional}, CodeBLEU-based consensus \cite{ren2020codebleu}, embedding-based consensus, and LLM-as-a-judge. Pass@1 measures expected single-sample accuracy, while Pass@$N$ measures whether at least one correct solution exists among the $N$ sampled candidates and therefore serves as an upper bound for selection. CodeT and SRank use generated tests to induce execution-based groups, while CodeBLEU and embedding consensus form similarity-based clusters from surface or representation features. LLM-as-a-judge prompts the generator model to select the best candidate among its own samples. Implementation details and the unbiased Pass@$k$ estimator are provided in Appendix~\ref{app:baselines}.

\subsection{Main Results}
\label{sec:results}

\begin{table}[t]
    \centering
\caption{\textbf{HumanEval+ Baselines vs Ours ($N=10$)} Comparison of selection strategies on the HumanEval+ dataset. The best performing non-oracle method among \textit{CodeBLEU}, \textit{Embedding}, \textit{LLM-Judge}, \textit{CodeT}, \textit{SRank}, and \textit{SEP (Ours)} for each model is \textbf{bold faced}.}
    \label{tab:heplus_baselines_ours_n10_all_models}
    \small
    \setlength{\tabcolsep}{3.2pt}
    \renewcommand{\arraystretch}{1.08}
    \makebox[\textwidth][c]{%
    \begin{tabular}{l|c|ccccc|c|c}
        \toprule
        & & \multicolumn{5}{c|}{\textbf{Baselines}} &  & \\
        \textbf{Model} & \textbf{Pass@1} & \textbf{CodeBLEU} & \textbf{Embedding} & \textbf{LLM-Judge} & \textbf{CodeT} & \textbf{SRank} & \textbf{SEP} & \textbf{Pass@10} \\
        \midrule
        \texttt{Llama-3.1-8B-Instruct}         & 0.589 & 0.604 & 0.622 & 0.585 & 0.635 & 0.657 & \textbf{0.689} & 0.787 \\
        \texttt{Phi-4-mini-instruct}           & 0.647 & 0.683 & 0.683 & 0.677 & 0.724 & 0.730 & \textbf{0.793} & 0.854 \\
        \texttt{Phi-4-reasoning}               & 0.826 & 0.842 & 0.854 & 0.835 & 0.863 & 0.865 & \textbf{0.902} & 0.927 \\
        \texttt{Qwen3-4B-Instruct-2507}        & 0.804 & 0.811 & 0.811 & \textbf{0.829} & 0.809 & 0.821 & \textbf{0.829} & 0.860 \\
        \texttt{Qwen3-4B}                      & 0.749 & 0.768 & 0.744 & 0.756 & 0.786 & 0.792 & \textbf{0.799} & 0.835 \\
        \texttt{Qwen3-14B}                     & 0.845 & 0.854 & 0.860 & 0.866 & 0.868 & 0.862 & \textbf{0.878} & 0.896 \\
        \texttt{gpt-oss-20b}                   & 0.820 & 0.817 & 0.860 & 0.780 & 0.864 & 0.858 & \textbf{0.890} & 0.951 \\
        \bottomrule
    \end{tabular}%
    }
\end{table}

\begin{table}[t]
    \centering
\caption{\textbf{LiveCodeBench Baselines vs Ours ($N=10$)} Comparison of selection strategies on the LiveCodeBench dataset. The best performing non-oracle method among \textit{CodeBLEU}, \textit{Embedding}, \textit{LLM-Judge}, \textit{CodeT}, \textit{SRank}, and \textit{SEP (Ours)} for each model is \textbf{bold faced}.}
    \label{tab:lcb_baselines_ours_n10_all_models}
    \small
    \setlength{\tabcolsep}{3.2pt}
    \renewcommand{\arraystretch}{1.08}
    \makebox[\textwidth][c]{%
    \begin{tabular}{l|c|ccccc|c|c}
        \toprule
        & & \multicolumn{5}{c|}{\textbf{Baselines}} &  & \\
        \textbf{Model} & \textbf{Pass@1} & \textbf{CodeBLEU} & \textbf{Embedding} & \textbf{LLM-Judge} & \textbf{CodeT} & \textbf{SRank} & \textbf{SEP} & \textbf{Pass@10} \\
        \midrule
        \texttt{Llama-3.1-8B-Instruct}         & 0.127 & 0.148 & 0.137 & 0.119 & 0.180 & 0.189 & \textbf{0.256} & 0.274 \\
        \texttt{Phi-4-mini-instruct}           & 0.169 & 0.185 & 0.178 & 0.176 & 0.213 & 0.217 & \textbf{0.256} & 0.290 \\
        \texttt{Phi-4-reasoning}               & 0.733 & 0.744 & 0.758 & 0.726 & 0.834 & 0.839 & \textbf{0.863} & 0.932 \\
        \texttt{Qwen3-4B-Instruct-2507}        & 0.521 & 0.534 & 0.543 & 0.557 & 0.579 & 0.572 & \textbf{0.621} & 0.692 \\
        \texttt{Qwen3-4B}                      & 0.749 & 0.744 & 0.772 & 0.731 & 0.759 & 0.757 & \textbf{0.795} & 0.854 \\
        \texttt{Qwen3-14B}                     & 0.830 & 0.820 & \textbf{0.858} & 0.831 & 0.832 & 0.832 & 0.854 & 0.922 \\
        \texttt{gpt-oss-20b}                   & 0.826 & 0.840 & 0.861 & 0.843 & 0.878 & \textbf{0.885} & 0.881 & 0.966 \\
        \bottomrule
    \end{tabular}%
    }
\end{table}

We evaluate Symbolic Equivalence Partitioning across seven models (4B to 20B parameters) at $N \in \{5,10,20\}$. Tables~\ref{tab:heplus_baselines_ours_n10_all_models} and \ref{tab:lcb_baselines_ours_n10_all_models} present the primary $N=10$ setting, and additional $N=5$, $N=20$ results are provided in Appendix~\ref{app:additional_results}. We treat $N=10$ as the default operating point because it balances selection quality and generation cost. At $N=10$, averaged across the seven models, SEP improves accuracy from 0.754 to 0.826 on HumanEval+ and from 0.565 to 0.647 on LiveCodeBench. Relative to the expanded baseline, SEP is best or tied-best in 40 of 42 evaluated settings.

\textbf{HumanEval+.} HumanEval+ does not expose domain constraints, so SEP is evaluated here without constraint injection. Even so, at $N=10$ SEP improves over Pass@1 on every model, with gains of +10.0 points on \texttt{Llama-3.1-8B-Instruct}, +14.6 on \texttt{Phi-4-mini-instruct}, and +7.6 on \texttt{Phi-4-reasoning}. SEP is best or tied-best among non-oracle methods for every model at $N=10$; on \texttt{Qwen3-4B-Instruct-2507}, it ties LLM-Judge at 0.829. Notably, this model also exhibits a very narrow gap between Pass@1 and oracle Pass@10, indicating limited room for improvement from reranking.

\textbf{LiveCodeBench.} At $N=10$, SEP is strongest on most models and remains close to the best baseline elsewhere, with large gains over Pass@1, including +12.9 points on \texttt{Llama-3.1-8B-Instruct} (0.127\,$\rightarrow$\,0.256), +8.7 on \texttt{Phi-4-mini-instruct} (0.169\,$\rightarrow$\,0.256), and +13.0 on \texttt{Phi-4-reasoning} (0.733\,$\rightarrow$\,0.863). The aggregate improvement is larger on LiveCodeBench than on HumanEval+ at $N=10$ (+8.2 vs.\ +7.2 points, or +14.5\% vs.\ +9.5\% relative), suggesting that SEP is especially useful on harder contest-style problems where syntactic similarity is less reliable than functional behavior.

\subsection{Additional Analysis}
\label{sec:analysis}

\paragraph{Scaling with candidate pool size.}
Figure~\ref{fig:scaling_plots} shows that Symbolic Equivalence Partitioning scales favorably with candidate pool size on both HumanEval+ and LiveCodeBench. As $N$ increases from 5 to 20, SEP achieves consistent improvements in average accuracy and maintains the strongest average performance among the compared methods. This indicates that SEP is already effective at $N=5$, making it practical even under tight candidate budgets, while continuing to improve through $N=20$. Taken together, these results suggest that symbolic equivalence partitioning can effectively exploit additional candidate diversity rather than saturating at small sample sizes.

\paragraph{Equivalence quality and failure modes.}

To assess the bounded-search approximation used in SEP, we first evaluate an operational pairwise equivalence accuracy: the agreement between SEP's bounded symbolic decision and a reference label induced by benchmark correctness, evaluated only on pairs containing at least one correct program (details in Appendix~\ref{app:pair_acc}). Accuracy remains high in aggregate (mean 0.884 over 42 model--budget settings), with means of 0.900 on HumanEval+ and 0.868 on LiveCodeBench. HumanEval+ remains strong even without constraint injection, while pairwise accuracy is somewhat lower on LiveCodeBench, likely reflecting deeper control flow and more complex algorithmic branching in contest-style problems.
We further analyze partition-level failure patterns on consensus-eligible instances, where at least two correct generations are present among the $N$ candidates, so a correct semantic consensus partition could exist. We measure two patterns: cases where the largest partition is mostly wrong, and cases where the largest partition contains at least one correct candidate but its representative is wrong. Across all seven models, both benchmarks, and all sampling budgets, 505 of 8,969 consensus-eligible instances (5.63\%) are flagged by at least one pattern. These low rates suggest that bounded symbolic partitioning usually preserves a useful consensus signal when multiple correct candidates are available, including on HumanEval+ where SEP runs without injected constraints, while the remaining failures are concentrated in identifiable partition-structure errors. Full model-level results are reported in Appendix~\ref{app:pair_acc}.

\begin{figure}[t]
    \centering
    \begin{minipage}{0.42\textwidth}
        \centering
        \begin{tikzpicture}
        \begin{axis}[
            width=\linewidth,
            height=4.2cm,
            xlabel={$N$},
            ylabel={Average accuracy},
            xmin=4, xmax=21,
            ymin=0.75, ymax=0.84,
            xtick={5,10,20},
            ymajorgrids=true,
            grid style=densely dotted,
            tick label style={font=\small},
            label style={font=\small},
            title style={font=\small},
            title={HumanEval+}
        ]
            \addplot+[blue, mark=o, thick] coordinates {
                (5,0.755) (10,0.754) (20,0.755)
            };

            \addplot+[red, mark=square*, thick] coordinates {
                (5,0.785) (10,0.793) (20,0.801)
            };

            \addplot+[green!60!black, mark=triangle*, thick] coordinates {
                (5,0.767) (10,0.768) (20,0.774)
            };

            \addplot+[orange!90!black, mark=diamond*, thick] coordinates {
                (5,0.767) (10,0.776) (20,0.771)
            };

            \addplot+[purple, mark=x, thick] coordinates {
                (5,0.762) (10,0.761) (20,0.759)
            };

            \addplot+[cyan!60!black, mark=pentagon*, thick] coordinates {
                (5,0.788) (10,0.798) (20,0.804)
            };

            \addplot+[black, mark=star, thick] coordinates {
                (5,0.808) (10,0.826) (20,0.827)
            };
        \end{axis}
        \end{tikzpicture}
    \end{minipage}
    \hfill
    \begin{minipage}{0.42\textwidth}
        \centering
        \begin{tikzpicture}
        \begin{axis}[
            width=\linewidth,
            height=4.2cm,
            xlabel={$N$},
            ylabel={Average accuracy},
            xmin=4, xmax=21,
            ymin=0.55, ymax=0.68,
            xtick={5,10,20},
            ymajorgrids=true,
            grid style=densely dotted,
            tick label style={font=\small},
            label style={font=\small},
            title style={font=\small},
            title={LiveCodeBench},
            legend to name=combinedlegend,
            legend columns=4,
            legend style={font=\scriptsize, /tikz/every even column/.append style={column sep=0.3em}},
        ]
            \addplot+[blue, mark=o, thick] coordinates {
                (5,0.566) (10,0.565) (20,0.565)
            };
            \addlegendentry{Pass@1}

            \addplot+[red, mark=square*, thick] coordinates {
                (5,0.605) (10,0.611) (20,0.614)
            };
            \addlegendentry{CodeT}

            \addplot+[green!60!black, mark=triangle*, thick] coordinates {
                (5,0.568) (10,0.574) (20,0.581)
            };
            \addlegendentry{CodeBLEU}

            \addplot+[orange!90!black, mark=diamond*, thick] coordinates {
                (5,0.581) (10,0.587) (20,0.587)
            };
            \addlegendentry{Embedding}

            \addplot+[purple, mark=x, thick] coordinates {
                (5,0.569) (10,0.569) (20,0.564)
            };
            \addlegendentry{LLM-Judge}

            \addplot+[cyan!60!black, mark=pentagon*, thick] coordinates {
                (5,0.607) (10,0.613) (20,0.619)
            };
            \addlegendentry{SRank}

            \addplot+[black, mark=star, thick] coordinates {
                (5,0.633) (10,0.647) (20,0.662)
            };
            \addlegendentry{Ours}
        \end{axis}
        \end{tikzpicture}
    \end{minipage}

    \vspace{0.3em}
    \ref{combinedlegend}

    \caption{\textbf{Scaling with candidate pool size.} Average accuracy across 7 models as the number of sampled candidates increases from $N=5$ to $N=20$ on HumanEval+ and LiveCodeBench. We compare Pass@1, five baselines, and our method.}
    \label{fig:scaling_plots}
\end{figure}

\paragraph{Constraint injection ablation.}
We ablate constraint injection (CI) on LiveCodeBench at $N=10$ using the same candidate pools. Constraint injection leaves representative accuracy nearly unchanged (0.647 vs. 0.646), but improves pairwise equivalence accuracy from 0.721 to 0.866 and reduces symbolic work by 17.4\%. Thus, CI mainly improves the robustness and efficiency of bounded symbolic comparison rather than changing the final selected representative. Full results are in Appendix~\ref{app:ci_ablation}.

\paragraph{Effect of public example filter.}
We ablate the public example filter by comparing SEP with filtering against SEP
without filtering at $N=10$. Removing the filter reduces representative accuracy
on both benchmarks, with average drops of 4.4 points on HumanEval+ and 4.0
points on LiveCodeBench. It also makes harmful dominant partitions more common:
mostly-wrong dominant partitions increase from 35 to 64 on HumanEval+ and from
78 to 151 on LiveCodeBench. The effect is especially visible on weaker
LiveCodeBench models, where more sampled generations fail public examples and
would otherwise enter the symbolic partitioning pool. These results support the
role of public examples as a lightweight grounding signal that removes
specification-violating candidates before symbolic partitioning. Full ablation
results are reported in Appendix~\ref{app:public-filter-ablation}.

\subsection{Runtime / Cost Analysis}
\label{sec:runtime}

Accuracy alone does not capture the practical cost of inference-time selection. SEP performs symbolic equivalence checking over the already generated candidate programs and requires no additional LLM calls beyond the initial $N$ candidate generations. In our implementation, this post-generation cost is CPU symbolic analysis on an AMD Ryzen 7 7700X. In contrast, test-generation baselines such as CodeT and SRank require additional H100 GPU inference to generate synthetic tests before using those tests for reranking. We report CodeT test-generation time as the representative cost because CodeT and SRank share the same generated tests in our implementation. Detailed accounting choices and per-model runtimes are reported in Appendix~\ref{app:runtime}.

SEP's runtime grows with the candidate pool size because larger $N$ requires more symbolic equivalence checks. Averaged across the seven models, SEP takes 539.4s, 1271.1s, and 2722.6s on HumanEval+ for $N=5,10,20$, respectively. On LiveCodeBench, it takes 1103.1s, 2583.9s, and 5291.9s. By contrast, CodeT's auxiliary test-generation cost is fixed with respect to $N$ because tests are generated once per problem before reranking the candidate pool; averaged across models, this cost is 11192.4s on HumanEval+ and 40986.0s on LiveCodeBench in our setup.

These measurements are intended as component-level resource accounting rather than a normalized CPU-vs-GPU speed comparison. Under this accounting, SEP's post-generation component has substantially lower measured wall-clock time while requiring no additional H100 inference.

\section{Limitations}
\label{sec:limitations}
While Symbolic Equivalence Partitioning performs strongly across our benchmarks, several limitations inherent to symbolic execution and our specific implementation remain.

\textbf{Language and Toolchain Dependence:}
Our current implementation relies on Python-specific symbolic execution engines. While the theoretical framework of symbolic consensus is language-agnostic, adapting it to other languages (e.g., C++, Java, Rust) requires robust symbolic execution backends (e.g., KLEE or CBMC), which may require additional engineering effort.

\textbf{Scalability Constraints:}
The computational cost of our method scales as $O(N' \cdot M)$, where $N'$ is the number of post-filter candidate solutions and $M$ is the number of distinct semantic partitions discovered under bounded symbolic analysis. For large candidate pools, this verification cost may become prohibitive for real-time applications. However, our empirical results show that strong performance is achieved at modest sample sizes ($N=10$), substantially reducing the need for large-scale sampling in practice.

\textbf{Bounded Symbolic Analysis \& Greedy Partitioning:}
SEP relies on bounded symbolic execution and representative-based greedy partitioning. Although operational pairwise equivalence accuracy is high in practice (Section~\ref{sec:analysis}), symbolic analysis is not exhaustive: if a semantic difference lies beyond the search horizon, the solver may fail to find a counterexample. Greedy partitioning further reduces comparison cost but can introduce order dependence, since candidates are compared only against current partition representatives. Our failure analysis suggests that problematic partition structures are uncommon: only 505 of 8,969 consensus-eligible instances (5.63\%) are flagged by either a mostly-wrong dominant partition or a largest partition containing a correct candidate but an incorrect representative. Still, these cases remain a source of error when bounded analysis merges non-equivalent programs or selects a wrong representative from a mixed partition.

\textbf{Dependence on Public Examples:}
The public example filter is an important grounding stage in SEP. When such examples are unavailable, SEP can still perform symbolic partitioning, but its selection signal relies more heavily on the assumption that the largest semantic partition corresponds to correct behavior. In these settings, correlated wrong solutions may be harder to eliminate.

\textbf{Problem Domain Constraints:}
This limitation is most relevant for competitive programming tasks, where programs are often highly sensitive to precise input constraints and boundary conditions. In such settings, constraint injection reduces exploration of invalid inputs by enforcing explicitly stated domain restrictions, improving the robustness and fidelity of pairwise symbolic comparisons. However, it does not reliably recover all constraints that are only implicitly specified in natural language or embedded in complex problem logic, so the symbolic engine may still explore unintended inputs, potentially causing spurious separation of semantically equivalent solutions. Nevertheless, SEP remains robust and achieves strong empirical results even when such constraints are not available.

\section{Conclusion}
\label{sec:conclusion}
We presented \textbf{Symbolic Equivalence Partitioning (SEP)}, an inference-time framework for selecting correct code solutions from multiple LLM generations. Unlike prior consensus-based methods that rely on auxiliary test generation or pretrained verifiers, SEP groups candidates by semantic behavior using symbolic execution.

Across HumanEval+ and LiveCodeBench, three sample budgets ($N=5,10,20$), and seven models, SEP consistently improves over \textit{Pass@1} and achieves the strongest accuracy among non-oracle baselines in 40 of 42 model--budget settings. In our component level resource accounting, SEP's CPU-side symbolic analysis has lower measured wall-clock time than CodeT's auxiliary H100 test-generation stage alone. These gains are achieved without requiring additional LLM inference beyond the initial $N$ candidate generations, both when domain constraints are available (LiveCodeBench) and when they are not (HumanEval+).

Although symbolic execution remains limited by bounded solver budgets, and SEP's robustness can depend on the availability of public examples and domain constraints, our results show that formal semantic analysis can serve as a practical and effective inference-time signal for code selection, complementing sampling-based generation without requiring auxiliary LLM inference.

\clearpage
\bibliography{templateArxiv}
\bibliographystyle{plainnat}

\newpage
\appendix
\onecolumn

\section{Additional Result Tables}
\label{app:additional_results}

\begin{table}[htbp]
    \centering
\caption{\textbf{HumanEval+ Baselines vs Ours ($N=5$).} Comparison of selection strategies on HumanEval+. Entries tied for best non-oracle performance among \textit{CodeBLEU}, \textit{Embedding}, \textit{LLM-Judge}, \textit{CodeT}, \textit{SRank}, and \textit{Ours} are \textbf{bolded}.}
    \label{tab:heplus_baselines_ours_n5_all_models}
    \small
    \setlength{\tabcolsep}{3.2pt}
    \renewcommand{\arraystretch}{1.08}
    \makebox[\textwidth][c]{%
    \begin{tabular}{l|c|ccccc|c|c}
        \toprule
        & & \multicolumn{5}{c|}{\textbf{Baselines}} &  & \\
        \textbf{Model} & \textbf{Pass@1} & \textbf{CodeBLEU} & \textbf{Embedding} & \textbf{LLM-Judge} & \textbf{CodeT} & \textbf{SRank} & \textbf{Ours} & \textbf{Pass@5} \\
        \midrule
        \texttt{Llama-3.1-8B-Instruct}         & 0.576 & 0.592 & 0.579 & 0.604 & 0.622 & 0.624 & \textbf{0.652} & 0.744 \\
        \texttt{Phi-4-mini-instruct}           & 0.656 & 0.659 & 0.665 & 0.677 & 0.715 & 0.713 & \textbf{0.750} & 0.823 \\
        \texttt{Phi-4-reasoning}               & 0.829 & 0.842 & 0.860 & 0.835 & 0.852 & 0.853 & \textbf{0.884} & 0.896 \\
        \texttt{Qwen3-4B-Instruct-2507}        & 0.804 & 0.817 & 0.817 & \textbf{0.823} & 0.811 & 0.817 & \textbf{0.823} & 0.854 \\
        \texttt{Qwen3-4B}                      & 0.748 & 0.762 & 0.756 & 0.744 & 0.773 & 0.782 & \textbf{0.793} & 0.829 \\
        \texttt{Qwen3-14B}                     & 0.848 & \textbf{0.872} & 0.866 & 0.854 & 0.861 & 0.861 & \textbf{0.872} & 0.884 \\
        \texttt{gpt-oss-20b}                   & 0.822 & 0.823 & 0.823 & 0.799 & 0.863 & 0.863 & \textbf{0.884} & 0.939 \\
        \bottomrule
    \end{tabular}%
    }
\end{table}

\begin{table}[htbp]
    \centering
\caption{\textbf{HumanEval+ Baselines vs Ours ($N=20$).} Comparison of selection strategies on HumanEval+. Entries tied for best non-oracle performance among \textit{CodeBLEU}, \textit{Embedding}, \textit{LLM-Judge}, \textit{CodeT}, \textit{SRank}, and \textit{Ours} are \textbf{bolded}.}
    \label{tab:heplus_baselines_ours_n20_all_models}
    \small
    \setlength{\tabcolsep}{3.2pt}
    \renewcommand{\arraystretch}{1.08}
    \makebox[\textwidth][c]{%
    \begin{tabular}{l|c|ccccc|c|c}
        \toprule
        & & \multicolumn{5}{c|}{\textbf{Baselines}} &  & \\
        \textbf{Model} & \textbf{Pass@1} & \textbf{CodeBLEU} & \textbf{Embedding} & \textbf{LLM-Judge} & \textbf{CodeT} & \textbf{SRank} & \textbf{Ours} & \textbf{Pass@20} \\
        \midrule
        \texttt{Llama-3.1-8B-Instruct}         & 0.587 & 0.628 & 0.616 & 0.598 & 0.664 & 0.682 & \textbf{0.701} & 0.823 \\
        \texttt{Phi-4-mini-instruct}           & 0.656 & 0.701 & 0.701 & 0.683 & 0.740 & 0.732 & \textbf{0.805} & 0.872 \\
        \texttt{Phi-4-reasoning}               & 0.830 & 0.860 & 0.866 & 0.829 & 0.865 & 0.867 & \textbf{0.896} & 0.945 \\
        \texttt{Qwen3-4B-Instruct-2507}        & 0.802 & 0.805 & 0.793 & 0.817 & 0.809 & 0.821 & \textbf{0.823} & 0.872 \\
        \texttt{Qwen3-4B}                      & 0.750 & 0.750 & 0.738 & 0.738 & 0.794 & 0.802 & \textbf{0.811} & 0.841 \\
        \texttt{Qwen3-14B}                     & 0.844 & 0.848 & 0.842 & 0.842 & 0.861 & 0.862 & \textbf{0.878} & 0.896 \\
        \texttt{gpt-oss-20b}                   & 0.816 & 0.829 & 0.842 & 0.805 & 0.873 & 0.862 & \textbf{0.878} & 0.957 \\
        \bottomrule
    \end{tabular}%
    }
\end{table}

\begin{table}[htbp]
    \centering
\caption{\textbf{LiveCodeBench Baselines vs Ours ($N=5$).} Comparison of selection strategies on LiveCodeBench. Entries tied for best non-oracle performance among \textit{CodeBLEU}, \textit{Embedding}, \textit{LLM-Judge}, \textit{CodeT}, \textit{SRank}, and \textit{Ours} are \textbf{bolded}.}
    \label{tab:lcb_baselines_ours_n5_all_models}
    \small
    \setlength{\tabcolsep}{3.2pt}
    \renewcommand{\arraystretch}{1.08}
    \makebox[\textwidth][c]{%
    \begin{tabular}{l|c|ccccc|c|c}
        \toprule
        & & \multicolumn{5}{c|}{\textbf{Baselines}} &  & \\
        \textbf{Model} & \textbf{Pass@1} & \textbf{CodeBLEU} & \textbf{Embedding} & \textbf{LLM-Judge} & \textbf{CodeT} & \textbf{SRank} & \textbf{Ours} & \textbf{Pass@5} \\
        \midrule
        \texttt{Llama-3.1-8B-Instruct}         & 0.126 & 0.139 & 0.132 & 0.121 & 0.180 & 0.178 & \textbf{0.221} & 0.231 \\
        \texttt{Phi-4-mini-instruct}           & 0.174 & 0.178 & 0.169 & 0.178 & 0.218 & 0.214 & \textbf{0.240} & 0.263 \\
        \texttt{Phi-4-reasoning}               & 0.738 & 0.744 & 0.779 & 0.740 & 0.831 & 0.831 & \textbf{0.854} & 0.900 \\
        \texttt{Qwen3-4B-Instruct-2507}        & 0.523 & 0.523 & 0.546 & 0.557 & 0.560 & 0.578 & \textbf{0.610} & 0.664 \\
        \texttt{Qwen3-4B}                      & 0.748 & 0.735 & 0.753 & 0.728 & 0.750 & 0.752 & \textbf{0.783} & 0.833 \\
        \texttt{Qwen3-14B}                     & 0.827 & 0.822 & 0.838 & 0.829 & 0.827 & 0.827 & \textbf{0.845} & 0.895 \\
        \texttt{gpt-oss-20b}                   & 0.827 & 0.836 & 0.849 & 0.831 & 0.870 & 0.872 & \textbf{0.879} & 0.936 \\
        \bottomrule
    \end{tabular}%
    }
\end{table}

\begin{table}[htbp]
    \centering
\caption{\textbf{LiveCodeBench Baselines vs Ours ($N=20$).} Comparison of selection strategies on LiveCodeBench. Entries tied for best non-oracle performance among \textit{CodeBLEU}, \textit{Embedding}, \textit{LLM-Judge}, \textit{CodeT}, \textit{SRank}, and \textit{Ours} are \textbf{bolded}.}
    \label{tab:lcb_baselines_ours_n20_all_models}
    \small
    \setlength{\tabcolsep}{3.2pt}
    \renewcommand{\arraystretch}{1.08}
    \makebox[\textwidth][c]{%
    \begin{tabular}{l|c|ccccc|c|c}
        \toprule
        & & \multicolumn{5}{c|}{\textbf{Baselines}} &  & \\
        \textbf{Model} & \textbf{Pass@1} & \textbf{CodeBLEU} & \textbf{Embedding} & \textbf{LLM-Judge} & \textbf{CodeT} & \textbf{SRank} & \textbf{Ours} & \textbf{Pass@20} \\
        \midrule
        \texttt{Llama-3.1-8B-Instruct}         & 0.128 & 0.151 & 0.158 & 0.121 & 0.186 & 0.202 & \textbf{0.292} & 0.324 \\
        \texttt{Phi-4-mini-instruct}           & 0.169 & 0.180 & 0.187 & 0.176 & 0.218 & 0.221 & \textbf{0.295} & 0.342 \\
        \texttt{Phi-4-reasoning}               & 0.734 & 0.758 & 0.760 & 0.737 & 0.840 & 0.857 & \textbf{0.874} & 0.952 \\
        \texttt{Qwen3-4B-Instruct-2507}        & 0.518 & 0.534 & 0.539 & 0.534 & 0.579 & 0.580 & \textbf{0.632} & 0.719 \\
        \texttt{Qwen3-4B}                      & 0.750 & 0.763 & 0.765 & 0.737 & 0.759 & 0.758 & \textbf{0.795} & 0.855 \\
        \texttt{Qwen3-14B}                     & 0.829 & 0.829 & 0.843 & 0.827 & 0.829 & 0.829 & \textbf{0.856} & 0.934 \\
        \texttt{gpt-oss-20b}                   & 0.825 & 0.849 & 0.854 & 0.817 & \textbf{0.888} & \textbf{0.888} & \textbf{0.888} & 0.982 \\
        \bottomrule
    \end{tabular}%
    }
\end{table}

\section{SEP Algorithm}
\label{app:sep_algorithm}

Algorithm~\ref{alg:partition} gives the full representative-based partitioning
procedure used by SEP. The algorithm first applies the public example filter to
the initial candidate pool, then greedily assigns each surviving candidate to an
existing partition if it is symbolically equivalent to that partition's
representative. Otherwise, it creates a new partition. The final prediction is
the representative of the largest partition.

\begin{algorithm}[tb]
\caption{Symbolic Equivalence Partitioning}
\label{alg:partition}
\begin{algorithmic}[1]
\STATE \textbf{Input:} Initial candidates $\mathcal{P}_0$, public examples $\mathcal{E}$, constraints $\mathcal{C}$
\STATE \textbf{Output:} Selected representative $p^*$

\STATE $\mathcal{P}_{\mathrm{f}} \gets \textsc{PublicExampleFilter}(\mathcal{P}_0, \mathcal{E})$
\STATE $\mathcal{K} \gets \emptyset$ \COMMENT{Initialize empty set of partitions}

\FOR{$p \in \mathcal{P}_{\mathrm{f}}$}
    \STATE $\textit{matched} \gets \text{false}$
    \STATE Sort $\mathcal{K}$ by cardinality $|K|$ descending \COMMENT{Check largest partitions first}

    \FOR{$K \in \mathcal{K}$}
        \STATE $r \gets \text{Representative}(K)$
        \IF{$\textsc{SymbolicCheck}(p, r, \mathcal{C})$ is \textsc{Equivalent}}
            \STATE $K \gets K \cup \{p\}$
            \STATE $\textit{matched} \gets \text{true}$
            \STATE \textbf{break}
        \ENDIF
    \ENDFOR

    \IF{\textbf{not} $\textit{matched}$}
        \STATE Create new partition $K_{\text{new}} \gets \{p\}$ with representative $p$
        \STATE $\mathcal{K} \gets \mathcal{K} \cup \{K_{\text{new}}\}$
    \ENDIF
\ENDFOR

\RETURN Representative of $\operatorname*{argmax}_{K \in \mathcal{K}} |K|$
\end{algorithmic}
\end{algorithm}

\section{Baseline Implementation Details}
\label{app:baselines}

To contextualize the performance of our selector, we define both lower and upper performance bounds using the unbiased $pass@k$ estimator \cite{chen2021evaluating}. Given $N$ generated samples where $c$ are functionally correct, the probability of selecting at least one correct solution in $k$ attempts is estimated as:
\begin{equation}
pass@k := 1 - \frac{\binom{N-c}{k}}{\binom{N}{k}}.
\end{equation}

We compare our \textbf{Symbolic Equivalence Partitioning} against the following baselines:

\begin{description}
    \item[Pass@1] The expected accuracy of a single, randomly selected solution ($pass@1 = c/N$). This represents the model's unguided performance.

    \item[Pass@N (The Oracle Ceiling)] The theoretical upper bound of the generator ($pass@N$). This represents the performance of an ideal selector that always retrieves a correct solution if at least one exists in the pool.

    \item[CodeT] A widely cited test-based reranking baseline for code generation. It uses test cases generated by the same code-generation model, executes each candidate program on those tests, and groups programs by identical execution outcomes. The resulting groups are ranked by a dual-agreement score that increases with both the number of generated tests passed by the group and the number of mutually agreeing candidate solutions it contains. The final prediction is then selected from the top-ranked group. Additional implementation details are included in Appendix~\ref{app:codet}.

    \item[SRank] A more recent execution-based reranking baseline introduced by \citet{to-etal-2024-functional} that extends test-based selection by modeling relationships between execution-induced solution clusters. Like prior work, it first uses generated test cases to obtain execution behavior for each candidate, then estimates functional overlap across clusters and uses these inter-cluster interactions to compute a final ranking over the candidate set. The top-ranked candidate is returned as the prediction. Compared with CodeT, SRank is designed to exploit higher-order structure among groups of functionally similar solutions rather than ranking clusters independently. In our implementation, we use the same generated test cases as CodeT to isolate differences in the reranking strategy.

    \item[Similarity-based clustering.] We evaluate two control baselines that cluster programs based on surface-level or latent features:
\begin{enumerate}
    \item \textbf{CodeBLEU Consensus:} We compute pairwise CodeBLEU scores \cite{ren2020codebleu} to capture syntactic and n-gram similarity.
    \item \textbf{Embedding Consensus:} We compute cosine similarity between dense vector representations generated by the \texttt{Qwen/Qwen3-Embedding-0.6B} model.
\end{enumerate}
In both cases, we apply Hierarchical Agglomerative Clustering (HAC) on the induced distance matrix to partition the candidate pool. The final output is selected as the \textit{medoid} of the largest cluster---defined as the candidate with the highest average similarity to all other members of that cluster.

    \item[LLM-as-a-Judge] A baseline where the generator model acts as a verifier to rerank its own $N$ candidates. The judge is prompted to select the single best candidate among the $N$ generations based on correctness and adherence to the specification. This serves as the most basic instantiation of verifier-based strategies.
\end{description}

\section{CodeT Implementation Details}
\label{app:codet}

For the \textbf{CodeT} baseline, we follow the dual-stage test-generation and execution framework described by \citet{chen2023codet} and use the official implementation provided at \url{https://github.com/microsoft/CodeT}.

An example test-generation prompt is shown below:
\begin{verbatim}
# Only output assert test cases.
# Use: assert function_signature(...) == ...

<starter_code>
pass

# check the correctness of <starter_code>
\end{verbatim}

CodeT proceeds in three stages:
\begin{enumerate}
    \item \textbf{Test Generation:} Following the original paper, we append a test-generation instruction $p$ to the starter code and request only executable \texttt{assert} statements, explicitly discouraging explanations or problem-solving logic. For each problem, we sample independent outputs and retain up to five valid \texttt{assert} statements that invoke the target entry point. We use 100 raw test-generation samples per problem on HumanEval+ (up to 500 generated tests). On LiveCodeBench, where longer problems and solutions make test generation substantially more expensive, we reduce the budget to 20 samples (up to 100 generated tests).
    \item \textbf{Execution:} All $N$ candidate solutions are executed against the resulting set of generated tests, producing pass/fail outcomes for each test.
    \item \textbf{Grouping and Ranking:} Solutions are represented by execution fingerprints---binary pass/fail vectors across tests---and grouped by identical fingerprints. Groups are ranked by a dual-agreement score (generated-test pass strength + agreement count), and the final answer is selected from the top-ranked group.
\end{enumerate}

The runtime measurements reported in Appendix~\ref{app:runtime} include only the auxiliary test-generation stage, which was the dominant overhead in our setup; candidate execution time is not included in those runtime tables.

\section{Sampling Parameters}
\label{app:param}

For all experiments, we generate candidate solutions using the default or recommended sampling configurations provided by each model's release documentation or generation configuration files. For models without explicit sampling recommendations, we use the same settings as a closely related model in the same family with similar size and training style. This choice ensures that generation quality reflects standard usage and avoids tuning sampling parameters to favor any particular selection method.

Table~\ref{tab:sampling_params} summarizes the temperature and top-$p$ values used for each evaluated model.

\begin{table}[htbp]
    \centering
    \caption{Sampling parameters used for candidate generation.}
    \label{tab:sampling_params}
    \small
    \begin{tabular}{lcc}
        \toprule
        \textbf{Model} & \textbf{Temperature} & \textbf{Top-$p$} \\
        \midrule
        \texttt{Phi-4-mini-instruct}      & 0.7 & 0.8  \\
        \texttt{Phi-4-reasoning}          & 0.8 & 0.95 \\
        \texttt{Qwen3-4B}                & 0.6 & 0.95 \\
        \texttt{Qwen3-4B-Instruct-2507}  & 0.7 & 0.8  \\
        \texttt{Qwen3-14B}               & 0.6 & 0.95 \\
        \texttt{Llama-3.1-8B-Instruct}    & 0.6 & 0.9  \\
        \texttt{gpt-oss-20b}             & 0.6 & 0.9  \\
        \bottomrule
    \end{tabular}
\end{table}

\section{Equivalence Quality and Failure Modes}
\label{app:pair_acc}

\subsection{Operational Pairwise Equivalence Accuracy}

We report an operational pairwise equivalence accuracy induced by benchmark correctness. The metric is computed only over candidate pairs containing at least one correct solution. Correct--correct pairs are treated as equivalent, correct--wrong pairs as non-equivalent, and wrong--wrong pairs are excluded because two incorrect programs may coincide on benchmark tests without being truly semantically equivalent, so benchmark failure alone does not determine semantic equivalence. This metric is not a complete semantic-equivalence benchmark; rather, it measures whether bounded symbolic comparison preserves agreement among correct solutions while separating correct solutions from incorrect ones.

Table~\ref{tab:pair_acc_heplus} and Table~\ref{tab:pair_acc_lcb} report full pairwise equivalence accuracy results for all models and sampling budgets.

\begin{table}[htbp]
    \centering
    \caption{Pairwise equivalence accuracy summary on HumanEval+ (higher is better).}
    \label{tab:pair_acc_heplus}
    \small
    \begin{tabular}{lcccc}
        \toprule
        \textbf{Model} & \textbf{Temp} & \textbf{$N=5$} & \textbf{$N=10$} & \textbf{$N=20$} \\
        \midrule
        \texttt{Llama-3.1-8B-Instruct}      & 0.6 & 0.853 & 0.852 & 0.847 \\
        \texttt{Phi-4-mini-instruct}        & 0.7 & 0.844 & 0.857 & 0.859 \\
        \texttt{Phi-4-reasoning}            & 0.8 & 0.876 & 0.877 & 0.879 \\
        \texttt{gpt-oss-20b}                & 0.6 & 0.824 & 0.833 & 0.830 \\
        \texttt{Qwen3-14B}                  & 0.6 & 0.954 & 0.951 & 0.954 \\
        \texttt{Qwen3-4B-Instruct-2507}     & 0.7 & 0.963 & 0.966 & 0.971 \\
        \texttt{Qwen3-4B}                   & 0.6 & 0.968 & 0.972 & 0.967 \\
        \bottomrule
    \end{tabular}
\end{table}

\begin{table}[htbp]
    \centering
    \caption{Pairwise equivalence accuracy summary on LiveCodeBench (higher is better).}
    \label{tab:pair_acc_lcb}
    \small
    \begin{tabular}{lcccc}
        \toprule
        \textbf{Model} & \textbf{Temp} & \textbf{$N=5$} & \textbf{$N=10$} & \textbf{$N=20$} \\
        \midrule
        \texttt{gpt-oss-20b}               & 0.6 & 0.831 & 0.817 & 0.813 \\
        \texttt{Llama-3.1-8B-Instruct}     & 0.6 & 0.901 & 0.866 & 0.846 \\
        \texttt{Phi-4-mini-instruct}       & 0.7 & 0.886 & 0.886 & 0.884 \\
        \texttt{Phi-4-reasoning}           & 0.8 & 0.834 & 0.828 & 0.818 \\
        \texttt{Qwen3-14B}                 & 0.6 & 0.874 & 0.874 & 0.876 \\
        \texttt{Qwen3-4B}                  & 0.6 & 0.888 & 0.880 & 0.884 \\
        \texttt{Qwen3-4B-Instruct-2507}    & 0.7 & 0.914 & 0.913 & 0.911 \\
        \bottomrule
    \end{tabular}
\end{table}

\subsection{Partition-Level Failure Pattern Breakdown}
\label{app:failure_modes}

We additionally report model-level partition failure patterns on consensus-eligible instances, where at least two correct generations are present among the $N$ sampled candidates. We consider two automatically measurable patterns. A \textit{mostly-wrong largest partition} indicates that the selected largest partition contains fewer correct than incorrect candidates. A \textit{wrong representative with correct member} indicates that the largest partition contains at least one correct candidate but its representative is incorrect. These patterns are not mutually exclusive; the ``Both'' column reports their overlap. Percentages are computed over consensus-eligible instances.

\begin{table}[htbp]
\centering
\caption{\textbf{HumanEval+ partition-level failure patterns by model.}
Results are aggregated over $N \in \{5,10,20\}$. Percentages are computed over consensus-eligible instances, with counts shown in parentheses. The two failure patterns are not mutually exclusive.}
\label{tab:failure_modes_heplus}
\small
\setlength{\tabcolsep}{3.0pt}
\begin{tabular}{lrrrrr}
\toprule
\textbf{Model}
& \textbf{Eligible}
& \textbf{Flagged}
& \textbf{Mostly-wrong}
& \textbf{Wrong rep.}
& \textbf{Both} \\
\midrule
\texttt{Llama-3.1-8B-Instruct}
& 349 & 9.17\% (32) & 6.59\% (23) & 4.30\% (15) & 1.72\% (6) \\
\texttt{Phi-4-mini-instruct}
& 387 & 5.43\% (21) & 4.91\% (19) & 1.81\% (7) & 1.29\% (5) \\
\texttt{Phi-4-reasoning}
& 441 & 2.49\% (11) & 2.49\% (11) & 0.23\% (1) & 0.23\% (1) \\
\texttt{Qwen3-4B-Instruct-2507}
& 416 & 3.12\% (13) & 3.12\% (13) & 0.48\% (2) & 0.48\% (2) \\
\texttt{Qwen3-4B}
& 400 & 3.00\% (12) & 2.75\% (11) & 1.25\% (5) & 1.00\% (4) \\
\texttt{Qwen3-14B}
& 431 & 1.86\% (8) & 1.86\% (8) & 0.00\% (0) & 0.00\% (0) \\
\texttt{gpt-oss-20b}
& 457 & 6.13\% (28) & 4.38\% (20) & 3.28\% (15) & 1.53\% (7) \\
\midrule
\textbf{Total}
& 2,881 & 4.34\% (125) & 3.64\% (105) & 1.56\% (45) & 0.87\% (25) \\
\bottomrule
\end{tabular}
\end{table}

\begin{table}[htbp]
\centering
\caption{\textbf{LiveCodeBench partition-level failure patterns by model.}
Results are aggregated over $N \in \{5,10,20\}$. Percentages are computed over consensus-eligible instances, with counts shown in parentheses. The two failure patterns are not mutually exclusive.}
\label{tab:failure_modes_lcb}
\small
\setlength{\tabcolsep}{3.0pt}
\begin{tabular}{lrrrrr}
\toprule
\textbf{Model}
& \textbf{Eligible}
& \textbf{Flagged}
& \textbf{Mostly-wrong}
& \textbf{Wrong rep.}
& \textbf{Both} \\
\midrule
\texttt{Llama-3.1-8B-Instruct}
& 277 & 3.25\% (9) & 1.44\% (4) & 2.17\% (6) & 0.36\% (1) \\
\texttt{Phi-4-mini-instruct}
& 328 & 9.15\% (30) & 6.10\% (20) & 5.18\% (17) & 2.13\% (7) \\
\texttt{Phi-4-reasoning}
& 1,164 & 6.10\% (71) & 3.35\% (39) & 4.47\% (52) & 1.72\% (20) \\
\texttt{Qwen3-4B-Instruct-2507}
& 824 & 6.43\% (53) & 4.98\% (41) & 3.64\% (30) & 2.18\% (18) \\
\texttt{Qwen3-4B}
& 1,078 & 4.73\% (51) & 2.78\% (30) & 3.71\% (40) & 1.76\% (19) \\
\texttt{Qwen3-14B}
& 1,191 & 7.14\% (85) & 4.28\% (51) & 4.70\% (56) & 1.85\% (22) \\
\texttt{gpt-oss-20b}
& 1,226 & 6.61\% (81) & 3.59\% (44) & 4.65\% (57) & 1.63\% (20) \\
\midrule
\textbf{Total}
& 6,088 & 6.24\% (380) & 3.76\% (229) & 4.24\% (258) & 1.76\% (107) \\
\bottomrule
\end{tabular}
\end{table}

\section{Constraint Injection Ablation}
\label{app:ci_ablation}

We ablate constraint injection (CI) on LiveCodeBench at $N=10$ using the same candidate pools as the main experiments. HumanEval+ is omitted because it does not provide domain constraints and is already evaluated without CI.

Table~\ref{tab:ci_ablation_per_model} reports per-model representative accuracy, pairwise equivalence accuracy, and symbolic work. Symbolic work is measured as the sum of per-problem symbolic-analysis time across all evaluated problems, making it less sensitive than script wall-clock time to worker scheduling, resume-based execution, and parallelism.

\begin{table}[htbp]
\centering
\caption{\textbf{Per-model constraint injection ablation on LiveCodeBench ($N=10$).}
CI improves pairwise equivalence accuracy and reduces symbolic work, while representative accuracy changes only slightly.}
\label{tab:ci_ablation_per_model}
\small
\setlength{\tabcolsep}{4pt}
\renewcommand{\arraystretch}{1.08}
\begin{tabular}{lcccccc}
\toprule
\textbf{Model} & \multicolumn{2}{c}{\textbf{Rep. Acc.}} & \multicolumn{2}{c}{\textbf{Pairwise Acc.}} & \multicolumn{2}{c}{\textbf{Symbolic Work (s)}} \\
\cmidrule(lr){2-3}
\cmidrule(lr){4-5}
\cmidrule(lr){6-7}
 & \textbf{w/ CI} & \textbf{w/o CI} & \textbf{w/ CI} & \textbf{w/o CI} & \textbf{w/ CI} & \textbf{w/o CI} \\
\midrule
\texttt{Qwen3-14B}              & 0.854 & 0.870 & 0.874 & 0.709 & 47236 & 57491 \\
\texttt{Qwen3-4B}               & 0.795 & 0.790 & 0.880 & 0.718 & 41572 & 49902 \\
\texttt{Qwen3-4B-Instruct-2507} & 0.621 & 0.621 & 0.913 & 0.812 & 27848 & 35090 \\
\texttt{gpt-oss-20b}            & 0.881 & 0.886 & 0.817 & 0.614 & 51421 & 62275 \\
\texttt{Llama-3.1-8B-Instruct}  & 0.256 & 0.251 & 0.866 & 0.803 & 6960  & 7600  \\
\texttt{Phi-4-mini-instruct}    & 0.256 & 0.256 & 0.886 & 0.783 & 10358 & 10667 \\
\texttt{Phi-4-reasoning}        & 0.863 & 0.849 & 0.828 & 0.610 & 43498 & 53928 \\
\midrule
\textbf{Average / Total}        & 0.647 & 0.646 & 0.866 & 0.721 & 228893 & 276953 \\
\bottomrule
\end{tabular}
\end{table}

Across all seven models, CI improves mean pairwise accuracy from 0.721 to 0.866 and reduces total symbolic work by 17.4\%, from 276953s to 228893s. Average representative accuracy changes only slightly, from 0.646 without CI to 0.647 with CI, indicating that CI primarily improves symbolic-comparison robustness and efficiency rather than changing the final selected representative.

\section{Public Example Filter Ablation}
\label{app:public-filter-ablation}

We ablate the public example filter at $N=10$ by comparing SEP with filtering
(WF) against SEP without filtering (NF). Representative accuracy is the final
accuracy of the candidate selected from the largest symbolic partition. A
mostly-wrong largest partition indicates that the selected largest partition
contains fewer correct than incorrect candidates. We report each paired value as
WF/NF.

\begin{table}[t]
\centering
\small
\setlength{\tabcolsep}{5pt}
\begin{tabular}{lccc}
\toprule
Model & Rep. Acc. & Mostly-wrong & Eligible \\
\midrule
\texttt{Qwen3-14B}
& 0.878 / 0.854 & 3 / 5 & 144 \\
\texttt{Qwen3-4B-Inst-2507}
& 0.829 / 0.817 & 4 / 7 & 140 \\
\texttt{Qwen3-4B}
& 0.799 / 0.756 & 5 / 11 & 136 \\
\texttt{Llama-3.1-8B-Inst}
& 0.689 / 0.622 & 8 / 11 & 119 \\
\texttt{Phi-4-mini-inst}
& 0.793 / 0.695 & 7 / 16 & 131 \\
\texttt{Phi-4-reasoning}
& 0.902 / 0.860 & 2 / 4 & 146 \\
\texttt{gpt-oss-20b}
& 0.890 / 0.866 & 6 / 10 & 154 \\
\bottomrule
\end{tabular}
\caption{Public example filter ablation on HumanEval+ at $N=10$. Each paired
entry is reported as with filtering / without filtering.}
\label{tab:public-filter-ablation-humaneval}
\end{table}

\begin{table}[t]
\centering
\small
\setlength{\tabcolsep}{5pt}
\begin{tabular}{lcc}
\toprule
Model & Rep. Acc. & Mostly-wrong \\
\midrule
\texttt{Qwen3-14B}
& 0.854 / 0.838 & 15 / 19 \\
\texttt{Qwen3-4B-Inst-2507}
& 0.621 / 0.566 & 16 / 36 \\
\texttt{Qwen3-4B}
& 0.795 / 0.769 & 7 / 12 \\
\texttt{Llama-3.1-8B-Inst}
& 0.256 / 0.178 & 0 / 16 \\
\texttt{Phi-4-mini-inst}
& 0.256 / 0.203 & 7 / 20 \\
\texttt{Phi-4-reasoning}
& 0.863 / 0.840 & 14 / 19 \\
\texttt{gpt-oss-20b}
& 0.881 / 0.854 & 19 / 29 \\
\bottomrule
\end{tabular}
\caption{Public example filter ablation on LiveCodeBench at $N=10$. Each paired
entry is reported as with filtering / without filtering.}
\label{tab:public-filter-ablation-lcb}
\end{table}

\section{Runtime Tables for SEP and CodeT}
\label{app:runtime}
We report CodeT runtime as the representative test-generation cost because CodeT and SRank use the same generated test cases in our implementation. Since auxiliary test generation is the dominant expense for these baselines, we report only CodeT's test-generation time and exclude subsequent candidate execution and ranking time. Therefore, the reported CodeT time should be interpreted as a lower bound on the full post-generation overhead of test-generation-based reranking.

\subsection{SEP Runtime (Total Wall-Clock Seconds)}

\begin{table}[htbp]
    \centering
    \caption{SEP runtime on HumanEval+ aggregated across the full benchmark evaluation.}
    \label{tab:sep_runtime_heplus}
    \small
    \begin{tabular}{lccc}
        \toprule
        \textbf{Model} & \textbf{$N=5$} & \textbf{$N=10$} & \textbf{$N=20$} \\
        \midrule
        \texttt{Qwen3-14B}                  & 612.44\,s & 1425.86\,s & 3000.88\,s \\
        \texttt{Qwen3-4B}                   & 589.46\,s & 1357.30\,s & 2853.48\,s \\
        \texttt{Qwen3-4B-Instruct-2507}     & 734.04\,s & 1659.71\,s & 3511.05\,s \\
        \texttt{Llama-3.1-8B-Instruct}      & 482.95\,s & 1189.17\,s & 2589.06\,s \\
        \texttt{Phi-4-mini-instruct}        & 665.81\,s & 1558.50\,s & 2891.10\,s \\
        \texttt{Phi-4-reasoning}            & 322.60\,s & 776.52\,s  & 1807.67\,s \\
        \texttt{gpt-oss-20b}                & 368.52\,s & 930.78\,s  & 2404.64\,s \\
        \bottomrule
    \end{tabular}
\end{table}

\begin{table}[htbp]
    \centering
    \caption{SEP runtime on LiveCodeBench aggregated across the full benchmark evaluation.}
    \label{tab:sep_runtime_lcb}
    \small
    \begin{tabular}{lccc}
        \toprule
        \textbf{Model} & \textbf{$N=5$} & \textbf{$N=10$} & \textbf{$N=20$} \\
        \midrule
        \texttt{Qwen3-14B}                  & 1634.37\,s & 3769.23\,s & 7822.59\,s \\
        \texttt{Qwen3-4B}                   & 1436.68\,s & 3317.21\,s & 6433.39\,s \\
        \texttt{Qwen3-4B-Instruct-2507}     & 973.82\,s  & 2222.15\,s & 4315.31\,s \\
        \texttt{Llama-3.1-8B-Instruct}      & 335.52\,s & 761.26\,s & 1719.09\,s \\
        \texttt{Phi-4-mini-instruct}        & 333.98\,s & 858.72\,s & 1852.35\,s \\
        \texttt{Phi-4-reasoning}            & 1479.12\,s & 3470.91\,s & 6519.91\,s \\
        \texttt{gpt-oss-20b}                & 1528.33\,s & 3688.03\,s & 8380.91\,s \\
        \bottomrule
    \end{tabular}
\end{table}

\subsection{CodeT Auxiliary Test-Generation Time}

\begin{table}[htbp]
    \centering
    \caption{CodeT auxiliary test-generation time on HumanEval+ aggregated across the full benchmark evaluation (format: HH:MM:SS). Candidate execution time is not included.}
    \label{tab:codet_runtime_heplus}
    \small
    \begin{tabular}{lc}
        \toprule
        \textbf{Model} & \textbf{Test-generation elapsed} \\
        \midrule
        \texttt{gpt-oss-20b}            & 00:50:12 \\
        \texttt{Llama-3.1-8B-Instruct}           & 00:14:13 \\
        \texttt{Phi-4-reasoning}                  & 00:22:15 \\
        \texttt{Phi-4-mini-instruct}             & 00:20:59 \\
        \texttt{Qwen3-14B}              & 10:52:58 \\
        \texttt{Qwen3-4B}               & 08:48:38 \\
        \texttt{Qwen3-4B-Instruct-2507} & 00:16:32 \\
        \bottomrule
    \end{tabular}
\end{table}

\begin{table}[htbp]
    \centering
    \caption{CodeT auxiliary test-generation time on LiveCodeBench aggregated across the full benchmark evaluation (format: HH:MM:SS). Candidate execution time is not included.}
    \label{tab:codet_runtime_lcb}
    \small
    \begin{tabular}{lc}
        \toprule
        \textbf{Model} & \textbf{Test-generation elapsed} \\
        \midrule
        \texttt{gpt-oss-20b}            & 20:41:55 \\
        \texttt{Llama-3.1-8B-Instruct}           & 04:11:43 \\
        \texttt{Phi-4-reasoning}                  & 16:13:01 \\
        \texttt{Phi-4-mini-instruct}             & 05:44:47 \\
        \texttt{Qwen3-14B}              & 15:10:39 \\
        \texttt{Qwen3-4B}               & 10:22:01 \\
        \texttt{Qwen3-4B-Instruct-2507} & 07:17:36 \\
        \bottomrule
    \end{tabular}
\end{table}

\section{Benchmark Details}
\label{app:benchmark_details}

\paragraph{HumanEval+ signature standardization.}
HumanEval+ prompts are heterogeneous in whether they include type-annotated
function signatures, while LiveCodeBench provides typed starter code for all
problems in our filtered subset. Because CrossHair uses Python type hints to
construct symbolic inputs, this difference can confound symbolic equivalence
checking with prompt formatting and with a model's tendency to emit type
annotations. To remove this confound, we standardize HumanEval+ candidates by
injecting the canonical entry-point signature from the EvalPlus contract for
problems whose prompts do not already provide a typed signature. Problems whose
prompts already contain typed signatures are left unchanged. The same
standardized candidate programs are used for SEP and all baselines.

\paragraph{Filtered LiveCodeBench subset.}

To ensure compatibility with function-level symbolic analysis and reliable evaluation, we excluded a subset of LiveCodeBench problems from our experiments. Starting from the full set of 1055 LiveCodeBench code-generation problems, we use the 438 LeetCode-style problems that provide starter code and callable function interfaces. We exclude 611 problems without starter code, consisting of 9 Codeforces and 602 AtCoder problems that require raw \texttt{stdin}/\texttt{stdout} parsing, and 6 additional problems with invalid or inconsistent ground-truth tests. No excluded problem was used in evaluation for any method reported in this paper. Exclusions fall into two categories:

\begin{itemize}
    \item \textbf{I/O-based problems.} Problems whose reference solutions rely on raw \texttt{stdin}/\texttt{stdout} parsing rather than a well-defined function signature were excluded.
    \item \textbf{Invalid or inconsistent test cases.} Problems for which the provided ground-truth test cases were found to be incorrect, incomplete, or internally inconsistent during manual inspection were excluded to avoid confounding evaluation results.
\end{itemize}

Table~\ref{tab:excluded_problems} lists the manually excluded problems with invalid or inconsistent test cases. I/O-based problems were excluded by the function-signature compatibility filter and are not listed individually.

\begin{table}[htbp]
    \centering
    \caption{Excluded LiveCodeBench problem titles.}
    \label{tab:excluded_problems}
    \small
    \begin{tabular}{l}
        \toprule
        \textbf{Problem Title} \\
        \midrule
        Minimum Average of Smallest and Largest Elements \\
        Apply Operations to Make String Empty \\
        Lexicographically Smallest String After Substring Operation \\
        Apply Operations to Make Two Strings Equal \\
        Find Maximum Removals from Source String \\
        Maximum Number of Groups with Increasing Length \\
        \bottomrule
    \end{tabular}
\end{table}

\section{Licenses of External Assets}
\label{app:licenses}

We summarize the licenses of the external datasets, models, and software artifacts used in this work. The entries below refer to the official repositories or model cards used to access the assets.

\subsection{Benchmarks and Software}

\begin{itemize}
    \item \textbf{HumanEval+} (used via \href{https://github.com/evalplus/evalplus}{EvalPlus}): the EvalPlus repository is released under the \textbf{Apache License 2.0}. Because HumanEval+ extends OpenAI \href{https://github.com/openai/human-eval}{HumanEval}, we also note that the original HumanEval repository is released under the \textbf{MIT License}.
    \item \textbf{LiveCodeBench}: the official \href{https://github.com/LiveCodeBench/LiveCodeBench}{LiveCodeBench repository} is released under the \textbf{MIT License}.
    \item \textbf{CrossHair}: the official \href{https://github.com/pschanely/CrossHair}{CrossHair repository} uses a mixed license: source files are \textbf{MIT}-licensed unless otherwise noted, and the repository also contains components distributed under the \textbf{Apache License 2.0} and the \textbf{PSF License Agreement}.
    \item \textbf{CodeT}: the official \href{https://github.com/microsoft/CodeT/tree/main/CodeT}{CodeT implementation} is released under the \textbf{MIT License}.
    \item \textbf{SRank}: the official \href{https://github.com/FSoft-AI4Code/SRank-CodeRanker}{SRank-CodeRanker repository} is released under the \textbf{MIT License}.
\end{itemize}

\subsection{LLMs Used in Experiments}

\begin{itemize}
    \item \texttt{Llama-3.1-8B-Instruct}: distributed under the \textbf{Llama 3.1 Community License Agreement}; see the official \href{https://huggingface.co/meta-llama/Llama-3.1-8B-Instruct}{model card}.
    \item \texttt{Phi-4-mini-instruct} and \texttt{Phi-4-reasoning}: the official Microsoft model cards list these checkpoints under the \textbf{MIT License} (\href{https://huggingface.co/microsoft/Phi-4-mini-instruct}{Phi-4-mini-instruct}, \href{https://huggingface.co/microsoft/Phi-4-reasoning}{Phi-4-reasoning}).
    \item \texttt{Qwen3-4B}, \texttt{Qwen3-4B-Instruct-2507}, and \texttt{Qwen3-14B}: the official Qwen model cards list these checkpoints under the \textbf{Apache License 2.0} (\href{https://huggingface.co/Qwen/Qwen3-4B}{Qwen3-4B}, \href{https://huggingface.co/Qwen/Qwen3-4B-Instruct-2507}{Qwen3-4B-Instruct-2507}, \href{https://huggingface.co/Qwen/Qwen3-14B}{Qwen3-14B}).
    \item \texttt{gpt-oss-20b}: the official OpenAI \href{https://huggingface.co/openai/gpt-oss-20b}{model card} lists this checkpoint under the \textbf{Apache License 2.0}.
\end{itemize}

\end{document}